\documentclass{article}
\usepackage{ijcai20}

\usepackage{fancyhdr}

\usepackage{times}

\usepackage{soul}
\usepackage{url}
\usepackage[hidelinks, colorlinks=true,% false: boxed links; true: colored links
    linkcolor=black,          % color of internal links (change box color with linkbordercolor)
    citecolor=black,        % color of links to bibliography
    filecolor=black,      % color of file links
    urlcolor=black]{hyperref}
\usepackage[utf8]{inputenc}
\usepackage[small]{caption}
\usepackage{graphicx}
\usepackage{amsmath}
\usepackage{booktabs}
\title{Flow-based Intrinsic Curiosity Module}

\makeatletter
\newcommand{\printfnsymbol}[1]{%
  \textsuperscript{\@fnsymbol{#1}}%
}
\makeatother

\author{
Hsuan-Kung Yang\thanks{Equal contribution}
\and
Po-Han Chiang\printfnsymbol{1}\and
Min-Fong Hong\And
Chun-Yi Lee
\affiliations
Elsa Lab, Department of Computer Science, National Tsing Hua University, Hsinchu, Taiwan\\
\emails
\{hellochick, ymmoy999, romulus, cylee\}@gapp.nthu.edu.tw}

\begin{document}

\maketitle
\thispagestyle{fancy} 

\pagestyle{fancy}

\fancyhf{} % sets both header and footer to nothing
\renewcommand{\headrulewidth}{0pt}

\fancyfoot[C]{The SOLE copyright holder is IJCAI (International Joint Conferences on Artificial Intelligence), all rights reserved.}

\begin{abstract}
In this paper, we focus on a prediction-based novelty estimation strategy upon the deep reinforcement learning (DRL) framework, and present a flow-based intrinsic curiosity module (FICM) to exploit the prediction errors from optical flow estimation as exploration bonuses.  We propose the concept of leveraging motion features captured between consecutive observations to evaluate the novelty of observations in an environment. FICM encourages a DRL agent to explore observations with unfamiliar motion features, and requires only two consecutive frames to obtain sufficient information when estimating the novelty.  We evaluate our method and compare it with a number of existing methods on multiple benchmark environments, including \textit{Atari} games, \textit{Super Mario Bros.}, and \textit{ViZDoom}.  We demonstrate that FICM is favorable to tasks or environments featuring moving objects, which allow FICM to utilize the motion features between consecutive observations.  We further ablatively analyze the encoding efficiency of FICM, and discuss its applicable domains comprehensively. See \href{https://github.com/hellochick/FICM}{\textcolor {blue}{\textbf{\textit{here}}}} for our codes and demo videos.
\end{abstract}

\section{Introduction}
A wide range of decision learning domains based on deep reinforcement learning (DRL) usually require an agent to explore environments containing moving objects, which are common in applications such as game playing~\cite{mnih2015human} and robot navigation~\cite{zhang2016learning}.
The rapidly changing parts or moving patterns between an agent's consecutive observations can be interpreted as an important indicator of information when exploring an environment.  A biological analogy is that animals often posses the capability to discover whether or not newcomers have intruded into their familiar territories. It is the motions of unfamiliar feature patterns appearing in an animal's field of view that arouses its  instinct of curiosity. Inspired by such a natural instinct, we argue that exploiting such motions features during the exploration phase might potentially add positive impacts to a DRL agent, especially for the above-mentioned domains.

\begin{figure}[t]
    \centering
    \includegraphics[width=.95\linewidth]{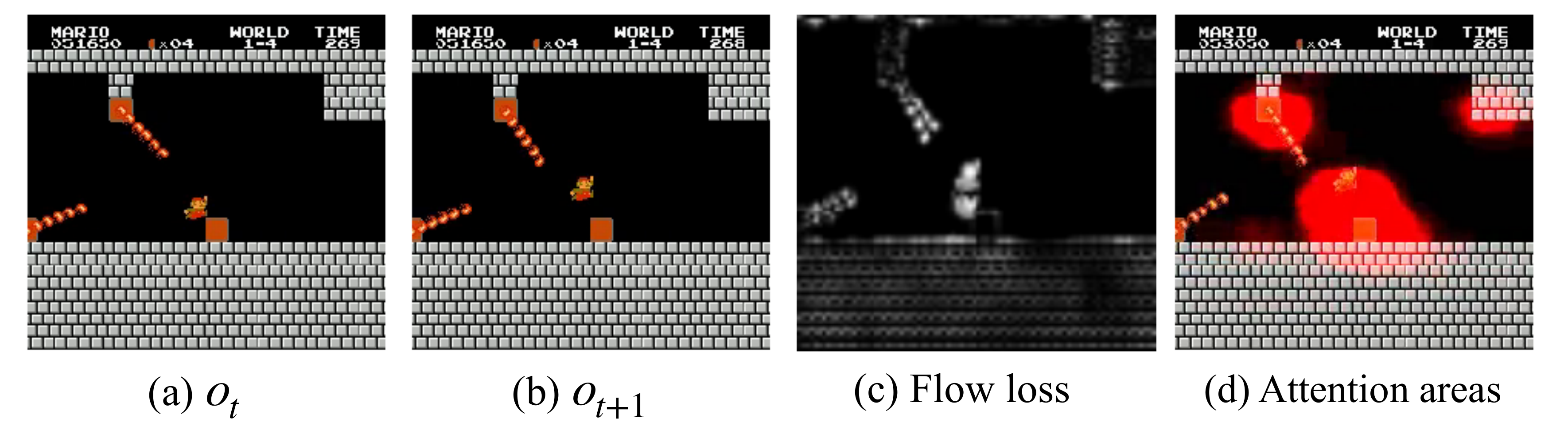}
    \caption{Illustration of the relation between the prediction errors (i.e., flow loss) and the attention areas of an agent. (a) and (b) are two consecutive frames $o_t$ and $o_{t+1}$. (c) shows the flow loss, while (d) depicts the attention areas \protect\cite{VisaulizeAtari} of the agent.}
    \label{fig:1}
\end{figure}

\begin{figure*}[t]
    \begin{minipage}[t]{.34\linewidth}
        \centering
        \includegraphics[width=.95\linewidth]{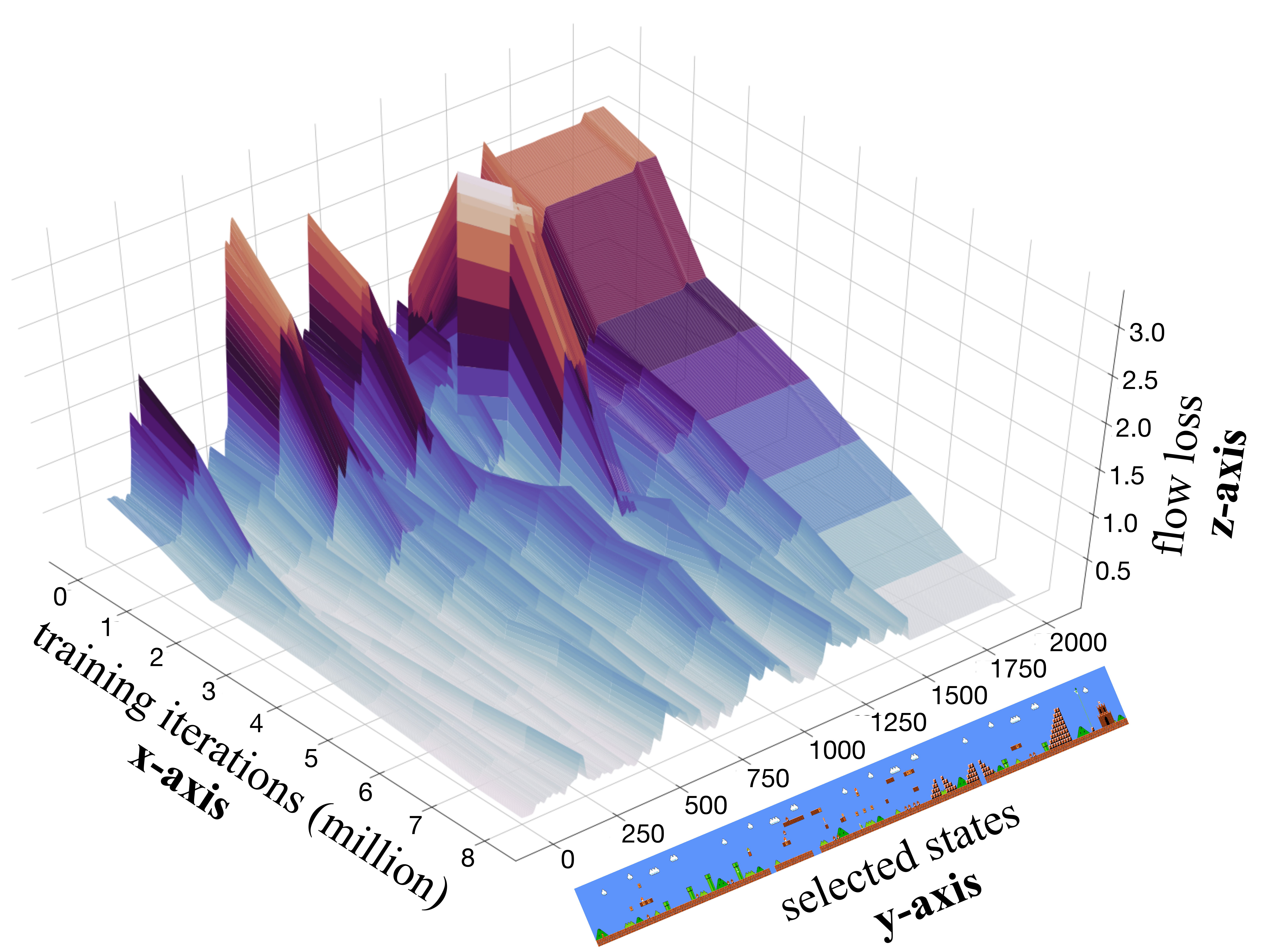}
        \caption{Prediction errors of FICM versus training iterations for selected states.}
        \label{fig:flowloss_time}
    \end{minipage}\hfill
    \begin{minipage}[t]{.64\linewidth}
        \centering
    \includegraphics[width=\textwidth]{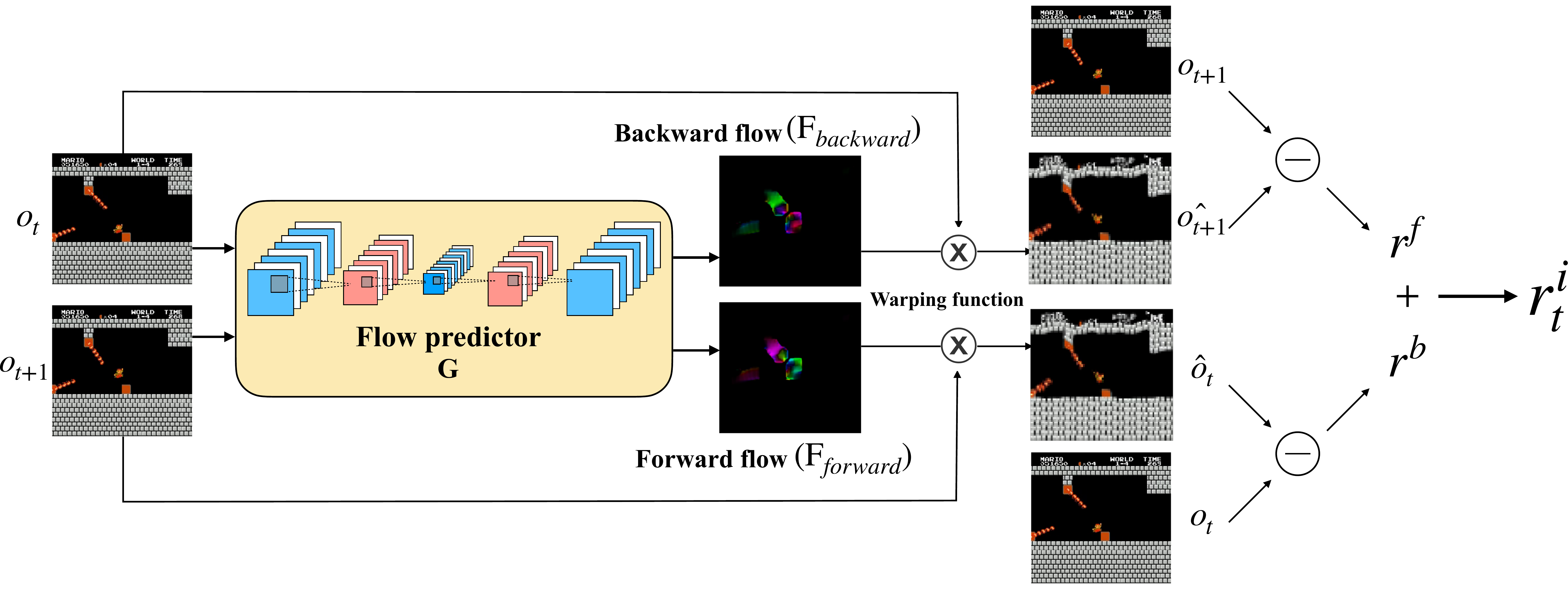}
    \caption{The workflow of the proposed flow-based intrinsic curiosity module (FICM).}
    \label{fig:workflow}
    \end{minipage}
\end{figure*}

Exploration incentives leveraging motion features captured between consecutive observations have not been discussed by the DRL community. Past researchers have attempted various ways to provide an agent with exploration bonuses (also known as ``\textit{intrinsic rewards}")  to encourage it to explore even when the reward signals from environments are sparse. These bonus rewards are associated with state novelty to incentivize an agent to explore novel states. A number of novelty measurement strategies have been proposed in the past few years, such as the use of information gain~\cite{houthooft2016vime}, count-based methods utilizing counting tables~\cite{bellemare2016exploration,CntExplore}, and prediction-based methods exploiting prediction errors of dynamics models~\cite{stadie2015incentivizing,ICM,LargeScale,RND}. The prediction-based novelty measurement strategies, which are the primary focus and interest of this paper, differ in the targets of prediction. Researchers in~\cite{ICM,LargeScale} introduce a forward dynamics model for predicting the feature representation of the next state of an agent, which is known as \textit{next-frame prediction}. Next-frame prediction for complex or rapid-changing motions, however, is rather difficult for forward dynamics models, especially when the prediction is made solely based on the current state and the taken action. It has been widely recognized that performing next-frame prediction typically requires complex feature representations~\cite{visual-dynamics,deep-pred-coding-net}. On the other hand, the researchers in \cite{RND} proposed a \textit{self-frame prediction} strategy by employing a random distillation network (RND) to predict the feature representation of the current state from a fixed and randomly initialized target neural network. Nevertheless, the attempt of self-frame prediction inherently neglects motion features between consecutive observations.

As a result, in this paper we introduce a \textbf{f}low-based \textbf{i}ntrinsic \textbf{c}uriosity \textbf{m}odule, called \textbf{FICM}, for evaluating the novelty of observations and generating intrinsic rewards. FICM is motivated by the issues mentioned above, and is designed to focus on the motion features of the objects extracted from two consecutive observations. FICM generates intrinsic rewards based on the prediction errors of optical flow estimation (i.e., the flow loss). Observations with low prediction errors (i.e., low intrinsic rewards) indicate that the agent has seen the observations plenty of times. On the contrary, observations are considered seldom visited when the errors from the predicted flow are non-negligible. The latter case then prompts FICM to generate high intrinsic rewards to encourage the agent for exploration. Fig.~\ref{fig:1} provides an illustrative example of FICM for \textit{Super Mario Bros}. It can be observed that the brighter parts (which correspond to higher flow loss) in Fig.~\ref{fig:1}~(c) align with the attention areas~\cite{VisaulizeAtari} in Fig.~\ref{fig:1}~(d), implying that the agent is intrinsically intrigued by the motion parts between the two observations.

We validate the above properties of FICM in a variety of benchmark environments, including \textit{Atari 2600}~\cite{bellemare2013arcade}, \textit{Super Mario Bros.}, and \textit{ViZDoom}~\cite{wydmuch2018vizdoom}. We demonstrate that FICM is preferable to a number of previous prediction-based exploration methods in terms of concentrating on motion features between consecutive observations in several tasks and environments. We further provide a set of ablation analyses for comparing the encoding efficiency of FICM and the existing approaches, and comprehensively discuss their implications. The primary contributions of this paper are thus summarized as follows. First, we propose FICM, which leverages on the prediction errors of optical flow estimation to encourage a DRL agent to explore unfamiliar motion features between consecutive observations in a comprehensive manner. Second, FICM is able to encode high dimensional inputs (e.g., RGB frames), and demands only two consecutive frames to obtain sufficient information when evaluating the novelty of the observations.

\section{Background}
\label{sec:background}
In this section, we first introduce the previous curiosity-driven exploration methodologies. We then discuss the two representative strategies employed by them: the next-frame prediction and the self-frame prediction strategies.  Finally, we briefly review the concepts of optical flow estimation.

\subsection{Curiosity-Driven Exploration Methodologies}
Curiosity-driven exploration is an exploration strategy adopted by a number of DRL researchers in recent years \cite{houthooft2016vime,bellemare2016exploration,CntExplore,ICM,LargeScale,RND} in order to explore environments more efficiently. While conventional random exploration strategies are easily trapped in local minima of state spaces for complex or sparse reward environments, curiosity-based methodologies tend to discover relatively un-visited regions, and therefore are likely to explore more effectively. In addition to the extrinsic rewards provided by the environments, most curiosity-driven exploration strategies introduce intrinsic rewards generated by the agent to encourage itself to explore novel states. In this paper, we primarily focus on the prediction-based curiosity-driven exploration methods, which include two branches: the next-frame and the self-frame prediction strategies.

\begin{figure*}[t]
    \centering
    \includegraphics[width=.8\textwidth]{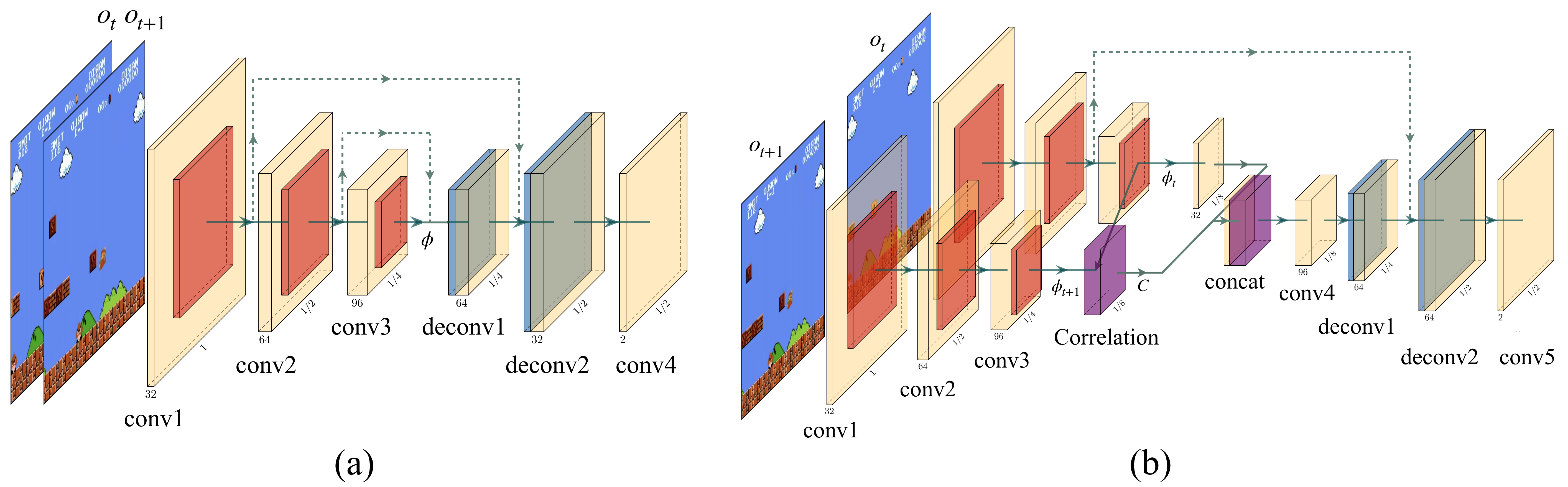}
    \captionsetup{width=.8\textwidth}
    \caption{The flow predictor architectures of (a) FICM-S and (b) FICM-C, which based on \textit{FlowNetS} and  \textit{FlowNetC} introduced by FlowNet 2.0~\protect\cite{FlowNet2}, respectively.}
    \label{fig:flow_archi}
\end{figure*}

\paragraph{Next-frame prediction strategy.} In \cite{ICM,LargeScale}, the authors formulate the exploration bonuses as the prediction errors of a forward dynamics network. Given an observation $o_t$, a feature representation $\phi_t$ is generated by an embedding network. The forward dynamics network $f$ is then used to predict the representation of the next observation by $\phi_t$ and the agent's taken action $a_t$. The mean-squared error (MSE) $||f(\phi_t, {a_t}) - \phi_{t+1})||^2$ is minimized by $f$ and serves as the intrinsic reward signal. When the same observations are visited multiple times, their MSE errors drop, indicating that the agent are becoming familiar with them. 

\paragraph{Self-frame prediction strategy.} In~\cite{RND}, the authors introduce RND to measure state novelty. RND is composed of two networks: a fixed and randomly initialized target network $g$, and a predictor network $\hat{g}$. For the current observation ${o_t}$, the objective of RND is to minimize the MSE $||g({o_t}) - \hat{g}({o_t})||^2$, which serves as the intrinsic reward signal. By introducing a deterministic $g$, RND avoids the problem of stochastic prediction errors occurred in next-frame prediction.

\subsection{Optical Flow Estimation}
Optical flow estimation is a technique for evaluating the motion of objects between consecutive images, which typically requires a reference image and a target image. Optical flow is usually represented as a vector field containing displacement vectors assigned to the pixels of the reference image. These vectors indicate the shifts of the corresponding pixels from the target image, and can be exploited as the motion features. In recent years, a number of deep learning approaches have been proposed~\cite{FlowNet,FlowNet2} for enhancing the prediction accuracy and resolving large displacement issues of optical flow estimation.

\section{Methodology}
\label{sec::methodology}
In this section, we present the motivation and design overview of FICM. We first introduce the essential concepts of FICM. Then, we formulate them into mathematical equations.  Finally, we investigate two different implementations of FICM.

\subsection{Flow-Based Curiosity-Driven Exploration}
\label{subsec::flow_based_curiosity_driven} 

Our main motivation and objective is to develop a new exploration bonus method that is able to (1) understand the motion features between two consecutive observations, (2) encode the observations efficiently, and (3) estimate the novelty of different observations. As a result, we propose to embrace optical flow estimation~\cite{FlowNet2,Unflow}, a technique that has commonly been leveraged by the field of computer vision for interpreting motion features as displacement vectors between consecutive frames, as our novelty measurement strategy. 
We validate the qualification of FICM to serve as a novelty estimator by an experiment performed on \textit{Super Mario Bros.}, and illustrate the decreasing trend of the flow loss for millions of iterations. Fig.~\ref{fig:flowloss_time} shows that FICM yields higher intrinsic rewards for unfamiliar observations than familiar ones. When the agent encounters unfamiliar observations, FICM motivates the agent to revisit these observations, and gradually learns the motion features of them. The workflow of FICM and the overview of the flow predictor design are illustrated in Figs.~\ref{fig:workflow} and~\ref{fig:flow_archi}, respectively.

\begin{figure*}[t]
    \centering
    \includegraphics[width=.8\linewidth]{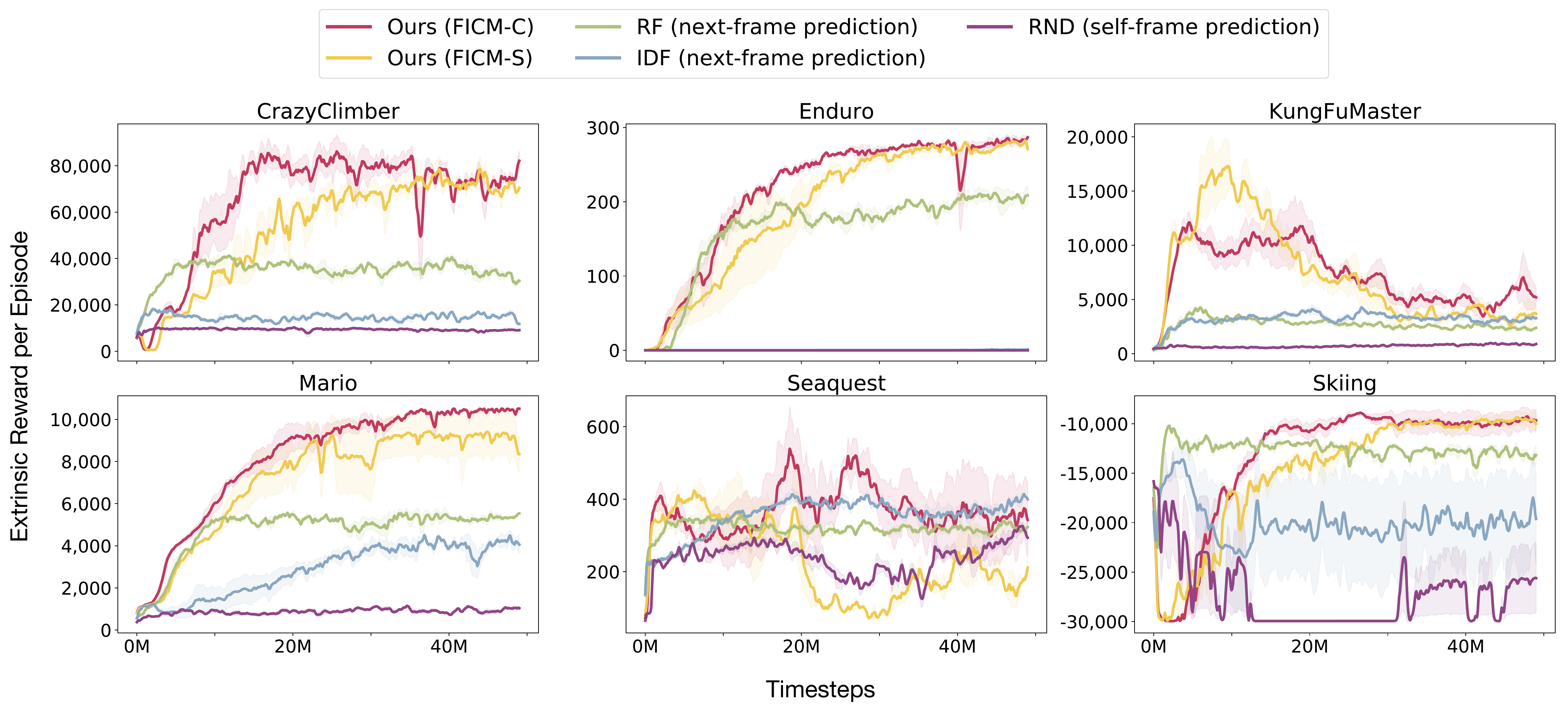}
    \caption{Mean extrinsic rewards (with standard errors) obtained by the agents fed only with intrinsic rewards during the training phase.}
    \label{fig:rewards_atari}
\end{figure*}

\subsection{Flow-Based Intrinsic Curiosity Module}
\label{subsec::FICM}
In this section, we formulate the procedure of FICM as formal mathematical equations.  The main objective of FICM is to leverage the optical flow between two consecutive observations as the encoded representation of them. 
Given two raw input observations $o_{t}$ and $o_{t+1}$ perceived at consecutive timesteps $t$ and $t+1$, FICM takes the 2-tuple $(o_{t}, o_{t+1})$ as its input, and predicts a forward flow $F_{forward}$ and a backward flow $F_{backward}$  by its flow predictor $G$ parameterized by a set of trainable parameters $\Theta_f$.  The two flows $F_{forward}$ and $F_{backward}$ can therefore be expressed as the following:

\begin{equation}
    \begin{aligned}
        &F_{forward} = G(o_{t}, o_{t+1}, \Theta_f)\\
        &F_{backward} = G(o_{t+1}, o_{t}, \Theta_f).\\
        ~
    \end{aligned}
\label{eq::1}
\end{equation}
$F_{forward}$ and $F_{backward}$ are  used to generate the observations $\hat{o_t}$ and $\hat{o_{t+1}}$ via a warping function defined in~\cite{FlowNet,FlowNet2}, where $\hat{o_t}$ and $\hat{o_{t+1}}$ are formulated as:

\begin{equation}
    \begin{aligned}
        &\hat{o_{t}}(x) = o_{t+1}(x + F_{forward}(x) * \beta)\\
        &\hat{o_{t+1}(x)} = o_{t}(x + F_{backward}(x) * \beta),\\
        ~
    \end{aligned}
\label{eq::2}
\end{equation}
where $\beta$ is a flow scaling factor, $x$ denotes the pixel index, and `$*$' represents element-wised multiplication. We use the bilinear interpolation method to implement the warping function. Please note that in this work, we employ the reverse mapping approach discussed in~\cite{thaddeus1992feature}, in which the warping procedure looks for the corresponding pixel coordinate in the source image for each pixel in the destination image. The forward mapping approach is not adopted in this work because some pixels in the destination image might not be found and correctly mapped from the source image.
%we employ inverse mapping, as the problem can be stated as: “which pixel coordinate in the source image do we sample for each pixel in the destination image?”~\cite{thaddeus1992feature}. 

With the predicted observations $\hat{o_{t}}$ and $\hat{o_{t+1}}$, $\Theta_f$ is iteratively updated to minimize the flow loss function $L_G$ of the flow predictor $G$, which consists of a forward flow loss $L^f$ and a backward flow loss $L^b$. The goal of $\Theta_f$ is expressed as:

\begin{equation}
    \begin{aligned}
        \min_{\Theta_f}L_G &= \min_{\Theta_f}(L^f + L^b)\\
        &= \min_{\Theta_f}({||o_{t+1} - \hat{o_{t+1}}||}^2 + {||o_{t} - \hat{o_{t}}||}^2),
    \end{aligned}
\label{eq::loss_function}
\end{equation}
where $(L^f, L^b)$ are derived from the MSE between $(o_{t+1}, \hat{o_{t+1}})$ and $(o_{t}, \hat{o_{t}})$, respectively.  In this work, $L_G$ is interpreted by FICM as a measure of novelty, and serves as the intrinsic reward signal $r^i$ presented to the agent.  The expression of $r^i$ can therefore be formulated as follows:

\begin{equation}
    \begin{aligned}
        r^i &= r^f + r^b = \frac{\zeta}{2}(L^f + L^b)\\
        &= \frac{\zeta}{2}L_G=\frac{\zeta}{2}({||o_{t+1} - \hat{o_{t+1}}||}^2 + {||o_{t} - \hat{o_{t}}||}^2),
    \end{aligned}
    \label{eq::intrinsic_reward}
\end{equation}
where $\zeta$ is the reward scaling factor, and $r^f$ and $r^b$ are the forward and backward intrinsic rewards scaled from $L^f$ and $L^b$, respectively.  Please note that $r^i$ is independent of the action taken by the agent, and FICM only takes two consecutive observations as inputs. The results presented in Section~\ref{sec::experiment} validate the effectiveness of the reward $r^i$ and our FICM.

\subsection{Implementations of FICM}
\label{subsec::implementation_FICM}
In this work, we propose two different implementations of FICM: FICM-S and FICM-C, as depicted in Fig.~\ref{fig:flow_archi} (a) and Fig.~\ref{fig:flow_archi} (b), respectively.
% These two implementations adopt different flow predictor architectures based on \textit{FlowNetS} and  \textit{FlowNetC} introduced by FlowNet 2.0~\cite{FlowNet2}, respectively.
We employ different implementations to validate that Eqs.~(\ref{eq::1})-(\ref{eq::intrinsic_reward}) are generalizable to different architectures, rather than restricted to any specific predictor designs. The primary difference between the two predictors lies in their feature extraction strategy from the input observations $o_t$ and $o_{t+1}$. FICM-S encodes the stacked observations $\left\langle o_{t}, o_{t+1}\right\rangle$ together to generate a single feature embedding $\phi$, while the other one encodes $o_t$ and $o_{t+1}$ in different paths to generate two separate feature embeddings $\phi_{t}$ and $\phi_{t+1}$.  In order to preserve both coarse, high-level information and fine, low-level information for enhancing the flow prediction accuracy, the feature embeddings in the two predictor designs are later fused with the feature maps from their shallower parts of the networks by skips~\cite{FlowNet}.  The fused feature map is then processed by another convolutional layer at the end to predict the optical flow from $o_t$ to $o_{t+1}$.  Please note that the two input paths of Fig.~\ref{fig:flow_archi} (b) are share-weighted in order to generate comparable embeddings.  The flow predictor of FICM-C additionally employs a correlation layer from~\cite{FlowNet} to perform multiplicative patch comparisons between $\phi_{t}$ and $\phi_{t+1}$. 

\begin{figure*}[t]
    \centering
    \includegraphics[width=.95\textwidth]{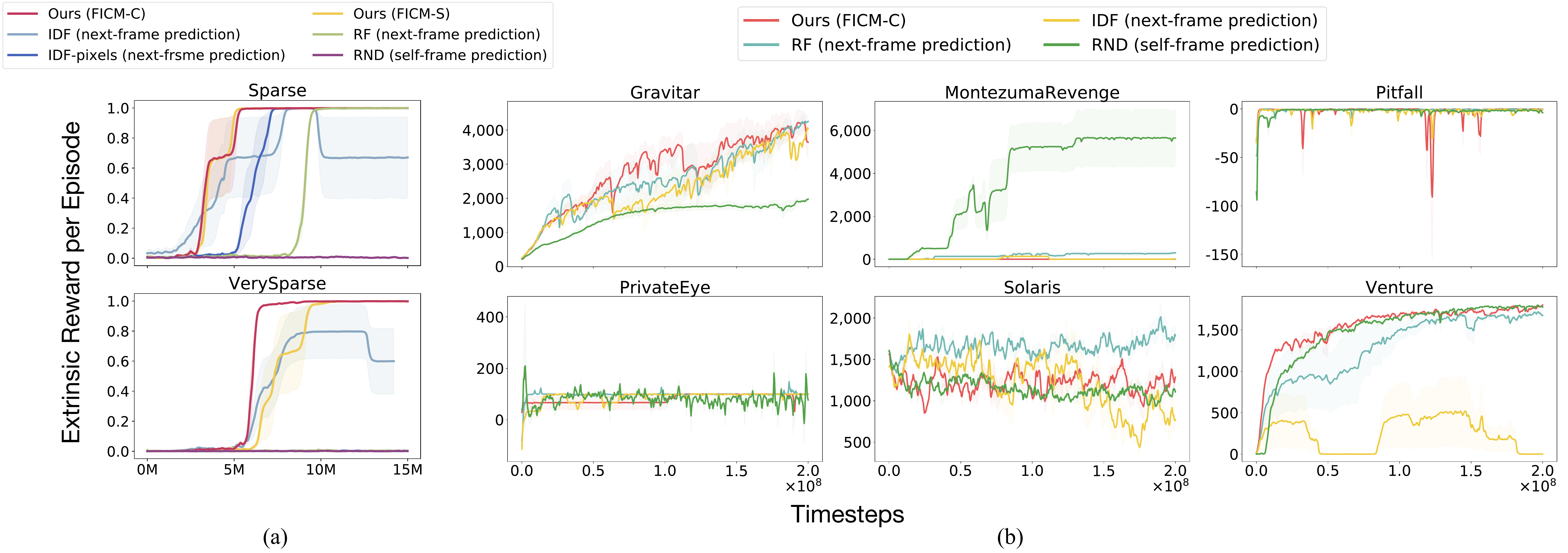}
    \caption{Comparison of the evaluation curves in (a)~\textit{ViZDoom} with \textit{sparse} and \textit{very sparse} extrinsic reward setups, and (b) six established hard exploration \textit{Atari} games.}
    \label{fig:rewards_doom}
\end{figure*}

\section{Experiments}
\label{sec::experiment}
In this section, we begin with a pure curiosity-driven learning task in the absence of any extrinsic reward signal in Section~\ref{subsec::no_reward}, to validate the exploitation capability of motion features. Then, we evaluate the performance of FICM on~\textit{VizDoom} as well as six established hard exploration~\textit{Atari} games, combining intrinsic rewards with sparse extrinsic rewards in Sections~\ref{subsec::sparse_reward} and~\ref{subsec::hard_explored}, respectively. In these experiments, we compare FICM against three baselines: next-frame prediction with (a) inverse dynamics features (\textit{IDF}) \cite{ICM} and (b) random features (\textit{RF}) \cite{LargeScale}, as well as (c) self-frame prediction with \textit{RND} \cite{RND}, respectively. To present the best settings in terms of the agents' performance, FICM takes RGB frames as the inputs of the flow predictor while the baselines maintain their default settings described in their original papers. An ablation analysis of the encoding efficiency between (a) RGB and gray-scale and (b) stack and non-stacked frames are further discussed in Section~\ref{subsec::ablation_analysis}. Finally, we discuss the impacts of FICM on the agents in Section~\ref{subsec::applicable_domains}. Unless stated otherwise, all of the experiments are conducted with proximal policy optimization (PPO)~\cite{PPO} and all of the presented curves are the average of three runs based on different random seeds.

\subsection{Experiments on FICM's Capability of Exploiting Motion Features for Exploration}
\label{subsec::no_reward}
\paragraph{Environments.}
We first validate FICM's capability of exploiting motion features to motivate agents for exploration by feeding the agents only with intrinsic rewards.  We similarly feed the baseline agents with only the intrinsic rewards specified in their original papers for a fair comparison.  In our experiments, we remove the end-of-episode signal, and measure the extrinsic rewards obtained by the agents during the training phase for~\textit{Super Mario Bros.} and five~\textit{Atari} games, including~\textit{CrazyClimber},~\textit{Enduro},~\textit{KungFuMaster},~\textit{Seaquest}, and~\textit{Skiing}.  These games are characterized by moving objects that require the agents to concentrate on and interact with.  More precisely, the motions of the foreground and background components (including the controllable agent and the uncontrollable objects) are related to the performance (i.e., obtainable scores) of the agent.  Taking~\textit{Enduro} for example, the agent not only has to understand the motions of its controllable car, but is also required to perceive and comprehend the motions of the other cars, as their motions are directly related to the final score of the agent.  We provide additional discussions on the above properties quantitatively in Section~\ref{subsec::applicable_domains}.

\paragraph{Results and discussions.} In Fig.~\ref{fig:rewards_atari},
we plot the evaluation curves of the extrinsic rewards obtained by the agents versus timesteps for our methods (denoted as~\textit{FICM-S} and~\textit{FICM-C}) and the baselines in different environments.  At each timesteps, different agents may focus their attention on exploring different subjects.  The resultant fluctuations in the curves (e.g.,~\textit{KungFuMaster} and~\textit{Seaquest}) directly reflect the pure exploration processes (and hence, the capability) of the agents, since there is no extrinsic reward provided to the them.  Please note that the same fluctuations in the curves also occurred in the experiments conducted in~\cite{LargeScale}.  The trends of the curves reveal that our methods enable the agents to explore states with high extrinsic rewards more often than the baselines in \textit{Super Mario Bros.} (e.g., discovering secret rooms and defeating Bowser), \textit{KungFuMaster}, \textit{CrazyClimber}, \textit{Enduro}, and \textit{Skiing}, while delivering comparable performance to the baselines in \textit{Seaquest}.  As these games are characterized by moving objects that require the agents to concentrate on and interact with, they are favorable to~\textit{FICM-S} and~\textit{FICM-C} that are capable of capturing motion features and perceiving the differences between observations.

\subsection{Experiments on Exploration with Sparse Extrinsic Rewards}
\label{subsec::sparse_reward}
\paragraph{Environments.}
We further conduct experiments on the \textit{DoomMyWayHome-v0} environment in \textit{ViZDoom}~\cite{wydmuch2018vizdoom}, using the same~\textit{sparse reward} and~\textit{very sparse reward} setups as those conducted in~\cite{ICM}. This environment features first-person perspective observations. Besides the default baselines mentioned in Section~\ref{sec::experiment}, we additionally include another next-frame prediction baseline \textit{IDF-pixels} (also known as \textit{ICM-pixels} in \cite{ICM}) for a fair comparison.  In these experiments, the DRL agents are all based on Asynchronous Advantage Actor-Critic (A3C)~\cite{mnih2016a3c}. As we intend to reproduce the results of~\cite{ICM} and compare with them, we adopt the officially released implementation of~\cite{ICM} as our baselines.  For a fair comparison, we modify the feature learning modules of~\cite{ICM} for the other baselines to generate different types of intrinsic rewards. Please note that in this environment, a single value head \cite{RND} is applied to~\textit{RND}.  The summation of the extrinsic and intrinsic rewards are provided to the DRL agents in all of the experiments considered in this section.

\paragraph{Results and discussions.}
Fig.~\ref{fig:rewards_doom}~(a) plots the evaluation curves of the experiments. It is observed that the agents based on FICM are able to converge faster than the baselines and maintain stable performance consistently in both the~\textit{sparse} and~\textit{very sparse} reward setups. The \textit{RF} and \textit{IDF-pixels} baselines are able to succeed in the \textit{sparse reward} setup, but fail in the \textit{very sparse reward} one. The \textit{IDF} baseline is able to reach the final destination in both cases, however, sometimes suffers from performance drops and is not always guaranteed to stably maintain its performance.  \textit{RND} fails to guide the agent to reach the final destination in both the \textit{sparse reward} and \textit{very sparse reward} setups. Since~\textit{FICM-C} is stabler than~\textit{FICM-S} in our results demonstrated in Sections~\ref{subsec::no_reward} and~\ref{subsec::sparse_reward}, we present only~\textit{FICM-C} in the remaining sections.

\subsection{Experiments on Hard Exploration Games}
\label{subsec::hard_explored}

\paragraph{Environments.} We evaluate FICM on six established hard exploration~\textit{Atari} games categorized by~\cite{bellemare2016exploration}, including~\textit{Venture},~\textit{Gravitar},~\textit{Montezuma's Revenge},~\textit{Pitfall},~\textit{PrivateEye}, and~\textit{Solaris}.  These games feature sparse reward environments.  In the experiments, the agents trained with the proposed methods and the baselines are fed with both extrinsic and intrinsic rewards. Please note that solving hard exploration games is not the focus of this paper. We provide these comparisons just for benchmark purposes, and validate if FICM is beneficial for sparse-reward games.

\paragraph{Results and discussions.} 
The evaluation curves are plotted and compared in Fig.~\ref{fig:rewards_doom}~(b). It is worth noticing that in~\cite{RND}, \textit{RND} established the state of art performance in~\textit{Montezuma's Revenge},~\textit{Gravitar}, and~\textit{Venture}. From Fig.~\ref{fig:rewards_doom}~(b), it can be observed that FICM outperforms the baselines in~\textit{Gravitar}, and is able to converge faster than them in~\textit{Venture}.  As the moving objects in these two games are related to the obtainable scores, focusing on motion features offers benefits for our agents to explore the environments. In~\textit{Solaris}, the scores of our method is comparable to those of~\textit{RND} and~\textit{IDF}, while~\textit{RF} achieves relatively higher performance. On the other hand, all of the methods in Fig.~\ref{fig:rewards_doom}~(b) achieve similar results in~\textit{Pitfall}, \textit{PrivateEye}, and~\textit{Montezuma's Revenge}, except for~\textit{RND} in~\textit{Montezuma's Revenge}.  The rationale is that these games demand reasoning and problem-solving skills, and are thus relatively harder to be solved solely by the observations perceived by the agents. 

\begin{figure}[t]
    \centering
    \includegraphics[width=\linewidth]{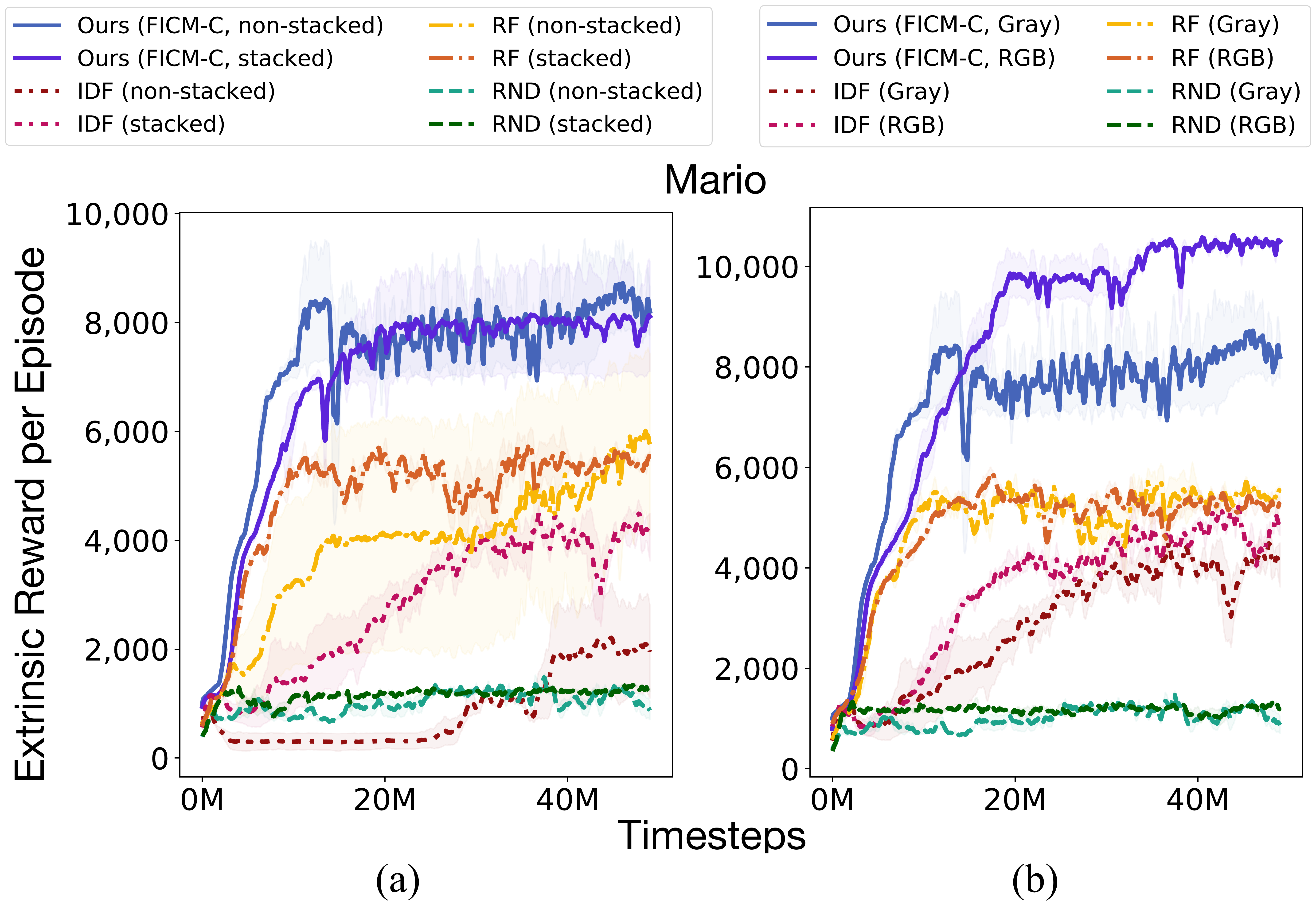}
    \caption{Encoding efficiency analysis: (a) stacked v.s. non-stacked frames and (b) RGB v.s. gray-scale. }
    \label{fig:encoding_efficiency}
\end{figure}

\subsection{Ablation Analysis}
\label{subsec::ablation_analysis}
In this section, we present a set of ablation analysis to investigate the feature encoding abilities of FICM in two aspects: (a) the number of the stacked frames and (b) the dimensionality of the input frames. The experiments are evaluated on \textit{Super Mario Bros.}, and the results are plotted in Fig.~\ref{fig:encoding_efficiency}, respectively. 

\paragraph{Stacked versus non-stacked frames.}
We consider two different setups in terms of the numbers of stacked frames: (a) two consecutive single frames (denoted as \textit{non-stacked}) and (b) a pair of four stacked frames (denotes as \textit{stacked}). The results show that both \textit{FICM-C (stacked)} and \textit{FICM-C (non-stacked)} maintain consistent performance, indicating that FICM is able to encode the information efficiently even with only two consecutive frames. In contrast, it is observed that the evaluation curves of \textit{RF (non-stacked)} and \textit{IDF (non-stacked)} exhibit high variance (highlighted as the wide shaded areas of the evaluation curves) and achieve lower scores as compared to their \textit{stacked} versions, indicating that they suffer from unstable performance due to the poor encoding efficiency. On the other hand, \textit{RND} shows similar encoding efficiency for the \textit{stacked} version and the \textit{non-stacked} version.

\paragraph{RGB versus gray-scale frames.}
We hypothesize that RGB frames contain more useful information than gray-scale ones to be utilized by prediction-based approaches. We examine whether FICM and the baselines are able to make use of the extra information to guide the agent for exploration. We present the experiments of utilizing RGB and gray-scale input frames for both FICM and the baselines in Fig.~\ref{fig:encoding_efficiency} (b). It is observed that \textit{RF} and \textit{RND} perform nearly the same for both settings. In contrast, FICM with RGB frames outperforms the one with gray-scale frames, indicating that FICM can encode the features more efficiently and utilize the information contained in RGB channels. The superior encoding efficiency is mainly due to the capability of the flow predictor in perceiving the motion features with color information, as it is much easier for it to distinguish between still and moving objects. It is observed that \textit{IDF} is also slightly benefited from the color information as compared to the one with gray-scale inputs. On the contrary, it is difficult for \textit{RF} and \textit{RND} to obtain appropriate representations of RGB frames, since they mainly rely on random features and are therefore lack of eligible embedding networks to encode the additional color information.

\begin{figure}[t]
    \centering
    \includegraphics[width=.80\linewidth]{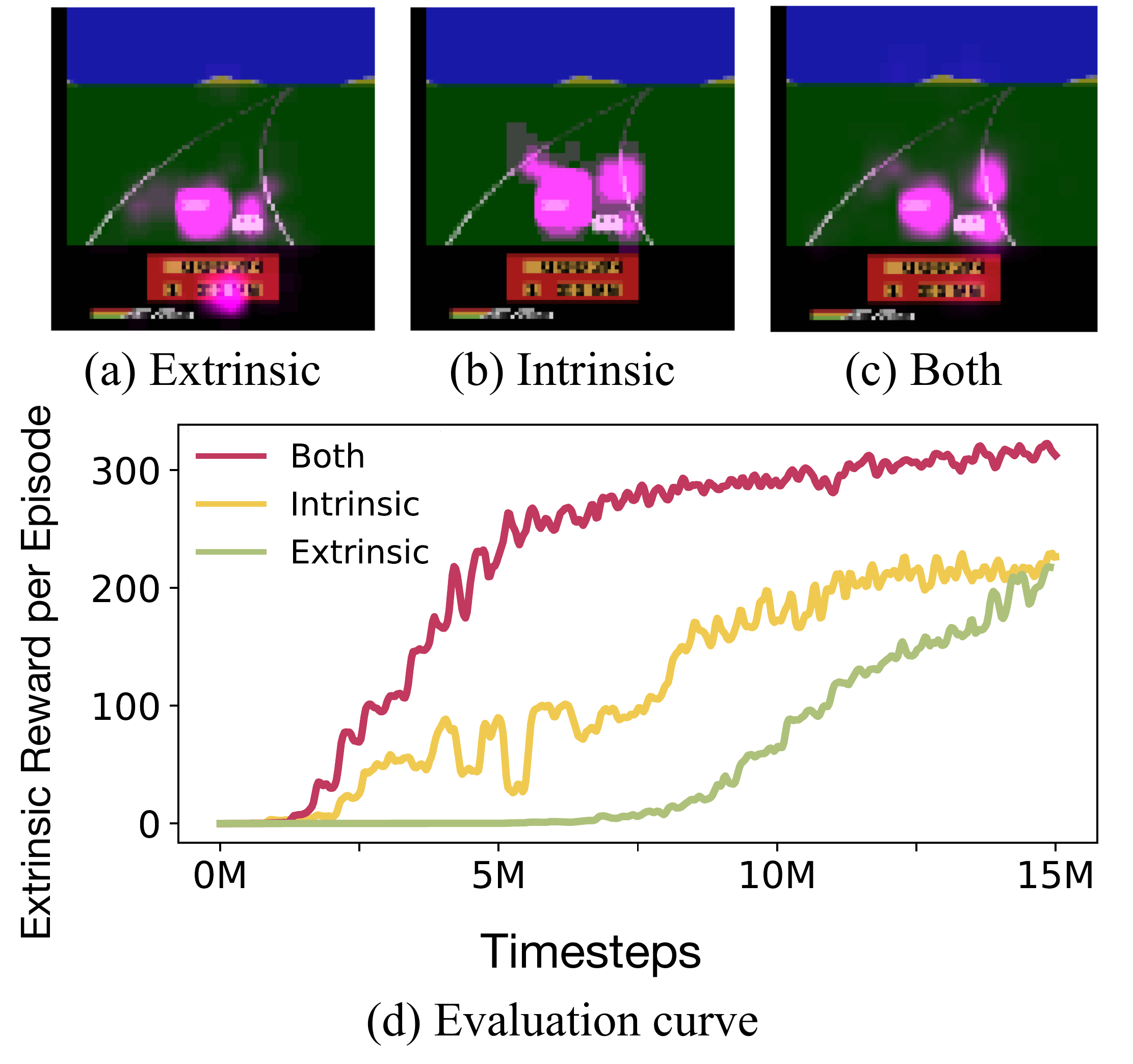}
    \caption{Visualization of the agents' attention areas (purple) and the evaluation curves for three reward setups.}
    \label{fig:explain_enduro}
\end{figure}

\begin{figure*}[t]
    \centering
    \includegraphics[width=.9\textwidth]{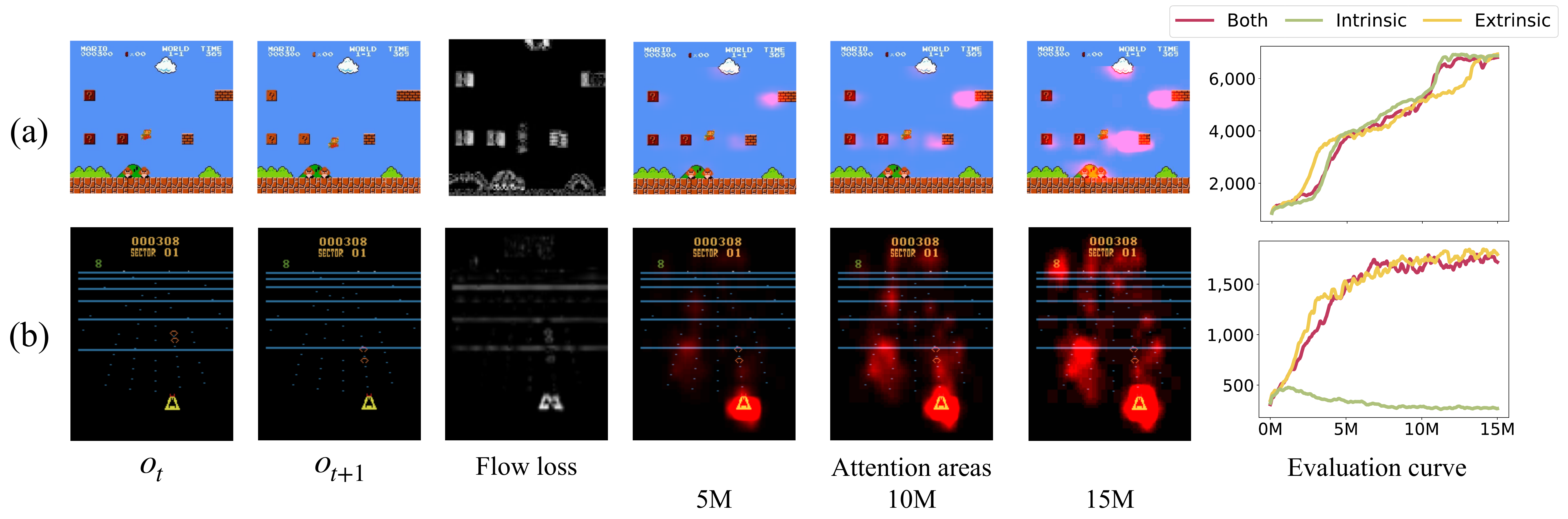}
    \caption{FICM's influence in (a) \textit{Super Mario Bros.} and (b) \textit{BeamRider}. The $1^{st}$-$3^{rd}$ columns show $o_t$, $o_{t+1}$, and the flow loss of them, where the brighter parts reflect higher flow loss. The $4^{th}$-$6^{th}$ columns highlight the agents' attention areas in red at different timesteps. The last one shows the evaluation curves.}
    \label{fig:explain_atari}
\end{figure*}

\subsection{Discussions on the Impacts of FICM}
\label{subsec::applicable_domains}
We first discuss why FICM is favorable for environments characterizing moving objects, which may include controllable agents and uncontrollable parts, by an illustrative example performed on \textit{Enduro}.  Fig.~\ref{fig:explain_enduro} visualizes the evaluation curves obtained with the same settings discussed in Section~\ref{subsec::no_reward} as well as the attention areas~\cite{VisaulizeAtari} for three different reward setups: (1) extrinsic rewards only, (2) intrinsic rewards only, and (3) both extrinsic and intrinsic rewards.  It is observed that in Fig.~\ref{fig:explain_enduro}~(a), the agent misfocuses its attention on the scoreboard at the bottom, which is not directly related to the performance of the agent. In contrast, Fig.~\ref{fig:explain_enduro}~(b) shows that the agent trained with FICM concentrates more on the moving parts (i.e., the controllable car and opponents). The evaluation curves in Fig.~\ref{fig:explain_enduro}~(d) also indicate that the agent trained with intrinsic rewards only is able to visit states with high extrinsic rewards more often than that trained with extrinsic rewards only. Fig.~\ref{fig:explain_enduro}~(d) further reveals that when applying both extrinsic and intrinsic rewards together during the train phase, the agent is able to perform significantly better than the other two cases.  It is observed that the attention areas depicted in Fig.~\ref{fig:explain_enduro}~(c) do not cover the scoreboard part anymore, implying that the intrinsic rewards generated by FICM bring positive impacts on guiding the learning process of the agent toward a beneficial direction.

Despite the benefits that FICM brings, FICM suits better for environments where the motions of the controllable and uncontrollable parts are relevant to the resultant performance.  FICM is not favorable in the following scenarios: (1) insufficiency of movements from the controllable and uncontrollable objects (e.g., enemies), and (2) excessive changes of irrelevant components (e.g., backgrounds) that might distract the focus of FICM.  For instance, if the objects that the agent is supposed to pay attention to barely move, FICM generates negligible intrinsic rewards due to little discrepancy between two consecutive observations.  On the other hand, if some irrelevant parts of the environment move relatively faster than those pertinent to the resultant performance, FICM may be distracted to focus on those irrelevant regions or components.

To prove and validate the hypothesis above, we further illustrate how FICM guides the agent to explore the environments with only intrinsic rewards by visualizing the corresponding flow loss and the attention areas~\cite{VisaulizeAtari} of the agents during the training phase in Fig.~\ref{fig:explain_atari}.  It is observed that the flow loss is highly correlated to the attention areas of the agents at different timesteps for both games. The evidence reveals that FICM is indeed able to guide the agents to concentrate on the regions with high flow loss during the training phase, as the attention areas grow wider and become brighter at those regions at later timesteps. Flow loss may as well distract the agent to focus on irrelevant parts of the agents' observations. Take \textit{BeamRider} for example, the agent is misled to pay attention to the background by the rich flow loss in it, limiting the extent of FICM's benefits. However, please note that the overall performance of the agents fed with both extrinsic and intrinsic rewards illustrated in the evaluation curves of Fig.~\ref{fig:explain_atari} are not degraded when they are compared with the agents fed with extrinsic rewards only, implying that the intrinsic rewards generated by FICM do not bring negative impacts to the agent's learning process. On the other hand, although the agent is able to master \textit{Super Mario Bros.} with different reward setups, the intrinsic rewards generated by FICM is able to support the agent's learning procedure to converge faster than that with only extrinsic rewards.  This is because FICM is able to help the agent to concentrate on the enemies, and not being distracted by the irrelevant objects. 

\section{Conclusions}
\label{sec::conclusion}
In this paper, we proposed FICM for evaluating the novelty of observations. FICM generates intrinsic rewards by leveraging on the prediction errors of optical flow estimation to encourage a DRL agent to explore unfamiliar motion features between consecutive observations in a comprehensive and efficient manner. We demonstrated the proposed methodology and compared it against a number of baselines on \textit{Atari} games, \textit{Super Mario Bros.}, and \textit{ViZDoom}. We validated that FICM is capable of focusing on the motion features and guiding the agents for exploration. Moreover, we presented a set of ablation analysis to investigate the encoding efficiency of FICM and discussed the impacts of it on the agents. Based on the experimental evidence discussed above, we thus conclude that FICM is a promising method to be employed by the agents, especially in environments featuring moving objects.

\section*{Acknowledgments}
% \section{Acknowlegment}
This work was supported by the Ministry of Science and Technology (MOST) in Taiwan under grant nos. MOST 109-2636-E-007-018 (Young Scholar Fellowship Program) and MOST 109-2634-F-007-017. The authors acknowledge the financial support from MediaTek Inc., Taiwan.  Hsuan-Kung Yang acknowledges the scholarship support from Novatek Microelectronics Corporation, Taiwan. The authors would also like to acknowledge the donation of the GPUs from NVIDIA Corporation and NVIDIA AI Technology Center.

% \bibliographystyle{named}
% \bibliography{ijcai20}

\end{document}